\documentclass[10pt,twocolumn,letterpaper]{article}

\usepackage[pagenumbers]{cvpr} 

\usepackage[dvipsnames]{xcolor}

\definecolor{cvprblue}{rgb}{0.21,0.49,0.74}
\usepackage[pagebackref,breaklinks,colorlinks,citecolor=cvprblue]{hyperref}

\def\paperID{998} 
\def\confName{CVPR}
\def\confYear{2024}

\definecolor{MyRed}{rgb}{1.0,0.0,0.0}

\title{MultiPLY: A Multisensory Object-Centric \\Embodied  Large Language Model in 3D World}

\author{Yining Hong$^{2,3}$, 
Zishuo Zheng$^{1}$,
Peihao Chen$^{1}$,
Yian Wang$^{1}$,
Junyan Li$^{1}$,
Chuang Gan$^{1,3}$\\
$^1$UMass Amherst, $^2$ UCLA, $^3$MIT-IBM Watson AI Lab \\\\
\url{https://vis-www.cs.umass.edu/multiply}
}

\begin{document}
\twocolumn[{
\renewcommand\twocolumn[1][]{#1}
\maketitle
\begin{center}

    \includegraphics[width=0.96\linewidth]{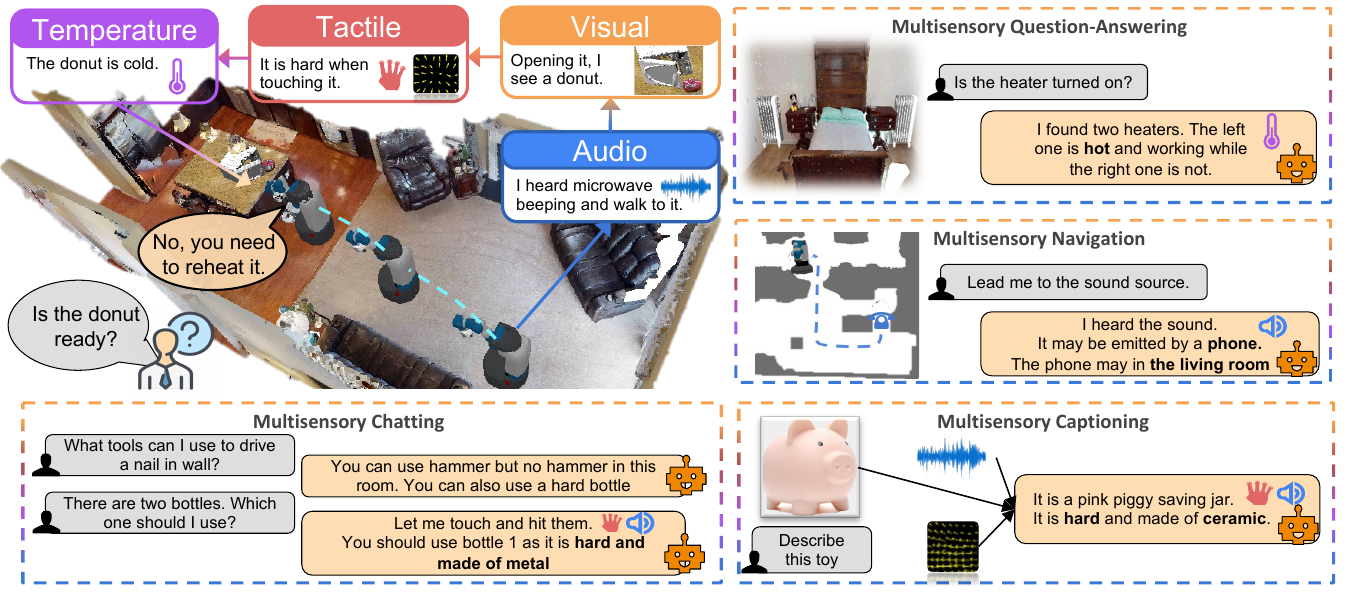}
    
    \captionof{figure}{We propose MultiPLY, a multisensory embodied LLM that encodes object-centric multisensory representations (\textit{e.g.,} visual, audio, tactile, and thermal), by deploying an embodied agent to engage with the 3D environment. MultiPLY excels at multiple tasks including multisensory captioning, question answering, dialogue, manipulation, navigation, tool use, task decomposition, and so on.}\label{fig:teaser}
\end{center}
}]
\begin{abstract} 
Human beings possess the capability to multiply a mélange of multisensory cues while actively exploring and interacting with the 3D world.
Current multi-modal large language models, however, passively absorb sensory data as inputs, 
lacking the capacity to actively interact with the objects in the 3D environment and 
dynamically collect their multisensory information. To usher in the study of this area,  we propose MultiPLY, a multisensory embodied large language model that could incorporate multisensory interactive data, including visual, audio, tactile, and thermal information into large language models, thereby establishing the correlation among words, actions, and percepts. To this end, we first collect Multisensory Universe, a large-scale multisensory interaction dataset comprising 500k data by deploying an LLM-powered embodied agent to engage with the 3D environment. To perform instruction tuning with pre-trained LLM on such generated data, we first encode the 3D scene as abstracted object-centric representations, and then introduce action tokens denoting that the embodied agent takes certain actions within the environment, as well as state tokens that represent the multisensory state observations of the agent at each time step. In the inference time, MultiPLY could generate action tokens, instructing the agent to take the action in the environment and obtain the next multisensory state observation. The observation is then appended back to the LLM via state tokens to generate subsequent text or action tokens. We demonstrate that MultiPLY outperforms baselines by a large margin through a diverse set of embodied tasks involving object retrieval, tool use, multisensory captioning, and task decomposition.

\end{abstract}    
\section{Introduction}
\label{sec:intro}

Human beings inhabit an extraordinary multisensory world - one in which we constantly explore and interact with the 3D environment, collecting and analyzing a mélange of sensory data to accomplish various tasks  \cite{Wallace2004TheDO}. 
Picture yourself situated within an embodied environment depicted as Figure \ref{fig:teaser}. To reason about the question ``is the donut ready for eating", you begin by hearing the microwave beep. Subsequently, you decide to investigate whether the donut is inside the microwave. Once you locate the donut, you may touch it, sensing its hardness and coldness, leading you to the conclusion that the donut is not yet ready.



Existing multi-modal large language models (\textit{e.g.,} LLaVA \cite{llava}, Flamingo \cite{alayrac2022flamingo}, BLIP-2 \cite{li2023blip2},  PaLM-E \cite{driess2023palme}) excel at numerous vision-language tasks. However, they mainly focus on 2D scene understanding, struggling to reason about and interact with 3D environments. Recent works such as 3D-LLM \cite{hong20233dllm} take preliminary steps to encode holistic 3D point clouds as inputs and show impressive results on 3D reasoning tasks, while suffering from expensive training and inefficient reasoning for objects.
More importantly, these models fall short of the ability to capture multisensory information that goes beyond vision and language. 

Efforts have been made to bind representations from different modalities \cite{girdhar2023imagebind}, and adapt them to pre-trained LLMs \cite{han2023imagebindllm, moon2023anymal}. However, they often focus on a single object \cite{guo2023pointbind} or 2D image \cite{girdhar2023imagebind}, unable to encode a large 3D environment and \textit{interact} with the 3D embodied environment. For example, to address a question illustrated in Figure \ref{fig:teaser}, a human would need to touch the donut to sense its softness and temperature, a capability well beyond the current scope of multi-modal LLMs.

Looking ahead, challenges inevitably exist for building embodied multisensory large language models. The first challenge resides in the paucity of multisensory interaction data for training such an LLM. 
The next challenge lies in the appropriate representations of the 3D scenes and multisensory information of the objects.
Humans could hold a coarse impression of the scene by abstracting the scene as an object-centric representation and attending to the object details when further interacting with the objects. It's essential for LLMs to also be able to flexibly switch between an abstracted object-centric representation and detailed multisensory information of the objects.
Lastly, existing LLMs are not tailored for instruction tuning with interaction data. They often take passive data as inputs and generate single-step outputs, incapable of connecting the words, actions, and percepts to engage with an embodied environment.

To this end, we propose MultiPLY, a multisensory embodied LLM that could encode multisensory object-centric representations, including visual, audio, tactile, and thermal information, by deploying an LLM-powered agent to engage with the 3D environment. We first collect Multisensory Universe, a large-scale multisensory dataset comprising 500k  data collected by an agent actively engaging with 3D embodied environments. We utilize the 3D environments from Habitat-Matterport 3D (HM3D) dataset \cite{ramakrishnan2021hm3d}, and enrich the environments by adding interactive objects with rich sensory data from ObjectFolder \cite{Gao2021ObjectFolderAD} and Objaverse \cite{deitke2022objaverse}. We prompt ChatGPT to create the input and output data of tasks ranging from multisensory captioning, question answering, dialogue, manipulation, task decomposition, and so on. An embodied agent explores the environment and interacts with the objects in the environment to get multisensory observations of these tasks. 

To perform instruction tuning on such generated data, we first encode the 3D scene as an abstracted object-centric representation, informing the LLM of what objects are in the scene. We further devise an additional set of action tokens such as \texttt{NAVIGATE}, \texttt{OBSERVE} (for obtaining object point cloud), \texttt{TOUCH} (for tactile and thermal information), \texttt{HIT} (for getting the impact sound) to denote that the agent takes the actions to explore the environment and interacts with the objects. By interacting with the objects, more detailed multisensory information could be unveiled as outcomes of the actions and encoded via a set of state tokens.
All sensory observations are encoded by different sensor encoders and connected to the LLM using sensor-to-image adapters.
 
In the inference time, MultiPLY could generate a series of action tokens through the LLM, instructing the agent to take the action and receive the outcome of the action as the next-state multisensory observation. The observation is then appended back to the LLM, enclosed by a set of state tokens, facilitating the next-step  generation. Our MultiPLY, trained on Multisensory Universe, outperforms baseline models by a large margin on object retrieval, tool use, multi-modal captioning, and task decomposition.

To sum up, the contributions of this paper are:
\begin{itemize}
\item We propose Multisensory Universe, a large-scale multisensory dataset comprising 500k data collected by an agent engaging with the 3D embodied environment, covering a diverse set of tasks involving multisensory captioning, question answering, dialogue, manipulation, task decomposition, and so on.
\item We propose MultiPLY, a multisensory embodied LLM that could encode multisensory object-centric representations with a novel set of action tokens and state tokens for the end-to-end instruction tuning of a pre-trained LLM.
\item Experimental results on object retrieval, tool use, multisensory captioning, and task decomposition show that MultiPLY outperforms baselines by a large margin.
\end{itemize}
\section{Related Works}

\textbf{Multisensory Learning}
Multisensory learning aims to learn from information from different sensors, including cameras, microphones, tactile sensors, etc. 
For visual-audio learning, the datasets collecting visual-audio pairs in real-world~\cite{clarke2023realimpact,owens2016visually} or rendering sounds in simulators~\cite{chen2020soundspaces,chen2022soundspaces2,gan2022finding} promote the development of this field of research.
Earlier works seek to combine audio and visuals information for audio-visual event localization~\cite{xu2020cross, xia2022cross, geng2023dense, wang2023context}, sound source localization in visual frame~\cite{Gan_2019_ICCV,gan2020music,zhao2019sound,Zhao_2018_ECCV,gao2018learning}, visual-guided sound editing~\cite{chen2022visual, garg2021geometry, gao2019visual-sound}, and visually-aligned sound generation~\cite{gan2020foley,chen2020generating,su2023physics,qi2023rd}.
As for visual-tactile learning, many works focus on building realistic tactile
simulation system~\cite{wang2022tacto,narang2021sim} or collecting tactile data of real objects ~\cite{gao2022ObjectFolderV2,gao2023objectfolder}. With these tactile data, researchers combine visual and tactile data for cross-modal retrieval~\cite{gao2021objectfolder,aytar2017see}, robotic manipulation~\cite{calandra2018more,calandra2017feeling,li2022see}, and 3D reconstruction~\cite{smith2021active,smith20203d,suresh2022shapemap}.
Different from the previous works, our MultiPLY aims to combine visual, audio, tactile, and thermal information in an interactive 3D environment for diverse embodied tasks.

\noindent\textbf{Multi-modal Large Language Models}
LLMs ~\cite{OpenAI2023GPT4TR,llama,opt,llama-2} demonstrate prowess across numerous domains. Recent works~\cite{alayrac2022flamingo,li2023blip2,llava} attempt to empower LLMs with visual understanding ability using large-scale image-text pair data and apply the trained models on downstream tasks like visual question-answering, image captioning, and multi-modal dialogue. Researchers~\cite{hong20233dllm, xu2023pointllm, sun20233d, yang2023llm} also focus on incorporating 3D visual information into LLMs to empower spatial reasoning abilities. In addition to incorporating visual information into LLMs, recent works~\cite{han2023imagebindllm,guo2023pointbind} attempt to enable LLMs to understand multi-modal information. AnyMAL~\cite{moon2023anymal} presents a unified model that aligns multi-modal information including text, image, video, audio, and IMU motion reading. 
However, these works process passive information rather than actively interact with the environment.
In contrast, our work focuses on an embodied large language model, which could actively interact with the multi-modal 3D world by navigating in the environment, touching objects to get tactile and thermal information, hitting objects to get impact sound, etc.

\section{The Multisensory-Universe Dataset}
\begin{figure}
    \centering
    \includegraphics[width=0.9\linewidth]{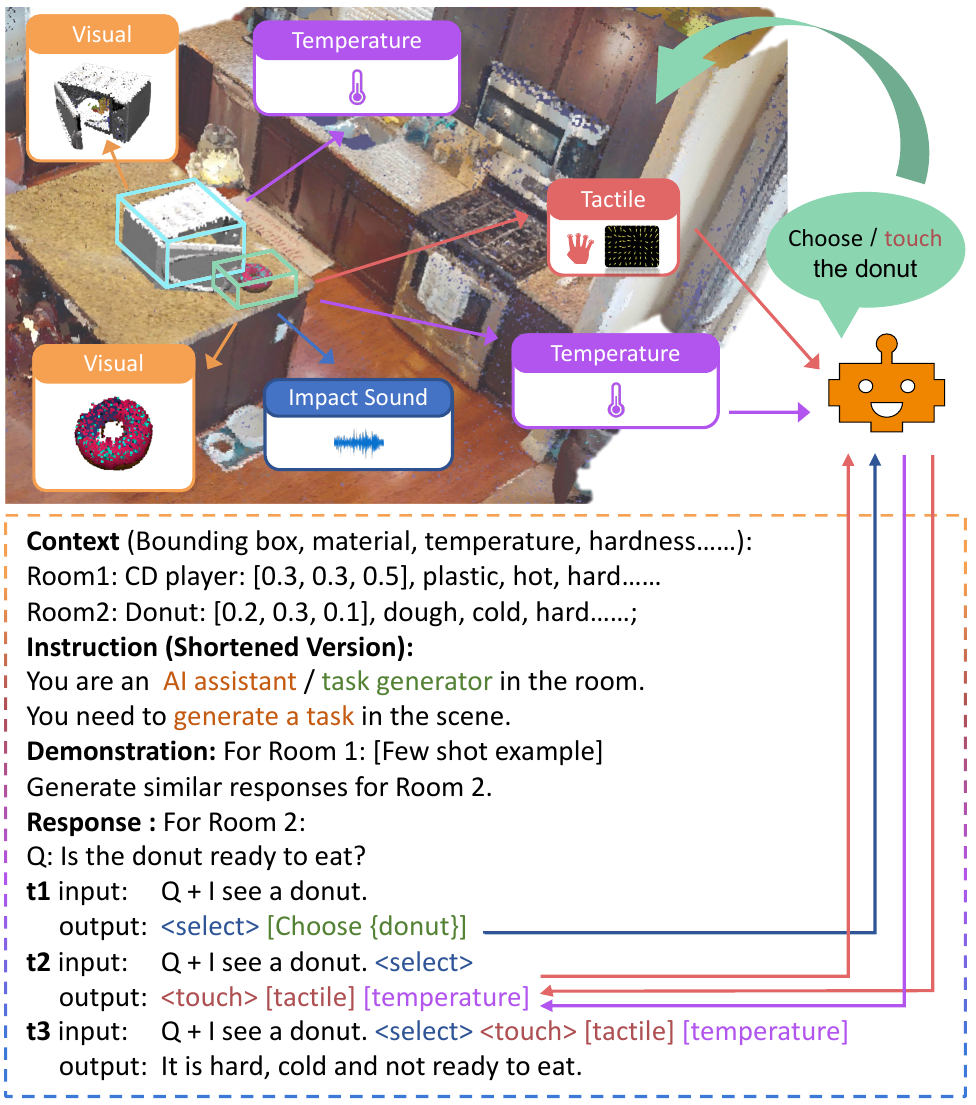}
    \vspace{-3mm}
    \caption{\textbf{Multisensory-Universe Generation Pipelines}. We first add a set of new interactive objects in the embodied environments, then prompt ChatGPT to generate diverse tasks about the environment. An embodied agent interacts with the objects to retrieve the multisensory information and construct interaction data. }
         \vspace{-4mm}
    \label{fig:dataset}
\end{figure}

In this section, we illustrate the process of collecting the Multisensory-Universe dataset. As presented in Figure \ref{fig:dataset}, we begin by explaining how we input interactive objects into the scene to construct object-centric 3D scenes for our dataset in Section \ref{sec:scene}. Subsequently, we outline the methodology for obtaining sensor data from these objects in Section \ref{sec:sensordata}. Moving on to Section \ref{sec: task}, we describe the deployment of an embodied agent tasked with proposing tasks and exploring the environment to solve them. The resulting interaction data are collected as paired interaction-language data, which serves as training input for the LLM.

\subsection{Inputting Interactive Objects into 3D Scenes}
\label{sec:scene}
We build our scenes on top of the Habitat-Matterport 3D (HM3D) semantics dataset \cite{ramakrishnan2021hm3d, yadav2022habitat}, which has 216 3D spaces and 3,100 rooms within those spaces. However, the existing objects in HM3D scenes, with insufficient sensor data and limited diversity, are not interactive in Habitat-sim \cite{habitat19iccv}. Thus, we propose to add new interactive objects to the scenes, allowing agents to interact with them using Habitat-sim.
The objects we add to the scenes are from two sources: 1) ObjectFolder \cite{Gao2021ObjectFolderAD, Gao2022ObjectFolder2A}, which contains 1k object meshes,  with impact sounds of these objects stored in implicit neural fields, and annotated with object materials. 2) Objaverse \cite{deitke2022objaverse} is a universe of 800K 3D objects spanning rich categories. We select the objects that could appear in indoor scenes.

We ask ChatGPT \cite{OpenAI2023GPT4TR} to choose 1-10 new objects from ObjectFolder and Objaverse, and generate the proper bounding boxes for these newly-added objects. ChatGPT is also required to specify objects' material categories (\textit{e.g.,} ceramic, plastic, steel) and properties(\textit{e.g.,}, deformation, elasticity hardness), as well as temperature labels (\textit{e.g.,} whether the objects are hot, cold, or the same as room temperature). 
Our prompt to GPT contains all existing objects in HM3D scenes and their bounding boxes, as well as several preferences: 1) Select some similar objects. For example, choose two bottles of similar appearances and specify one of them as plastic and the other one as steel. In this way, information from different sensors needs to be collected to resolve the ambiguity. 2) Select objects that are compatible with the environment and can be utilized together for interesting tasks. For instance, in a kitchen environment, we could put ingredients and tools for cooking. We also give some few-shot prompting examples to GPT.

\subsection{Object Sensor Data Acquisition}
\label{sec:sensordata}
We illustrate how we collect sensor data of added objects.
\begin{itemize}
\item \textbf{Tactile} We use DiffTactile \cite{anonymous2023difftactile} which leverages MLS-MPM \cite{Hu2018AML} to simulate rigid, elastic, elasto-plastic objects. We put meshes of added objects into DiffTactile, which uses the bubble gripper with several position markers to touch the objects at pre-defined positions. The tactile readings are the initial and final positions of the markers, which represent how much the bubble deforms.  


\item \textbf{Ambient Sound} 
Each object could emit ambient sound to facilitate navigation or reasoning, or serve as cues for informing the agents what's going on in the environment.
We prompt ChatGPT to match the sounds from AudioSet \cite{Gemmeke2017AudioSA} with the semantic labels of the added objects. Given the Audioset description, ChatGPT needs to select objects in the candidate list that are possible to make this sound.

\item \textbf{Impact Sound} Impact sound represents the sound that we hear when we strike or hit an object, which is crucial for identifying the material of an object. We get the impact sounds of ObjectFolder objects by querying their implicit sound fields given a hitting position and a force.

\item \textbf{Temperature} Given the temperature label of the object, we ask ChatGPT for a proper temperature of each object.
\end{itemize}


\subsection{Embodied Agents for Data Collection}
\label{sec: task}
Inspired by \cite{wang2023robogen}, we utilize LLM-powered embodied agents to collect the data in the constructed scenes. We first prompt ChatGPT to propose tasks. Then we place an embodied agent to interact with the objects in 3D environments to perform the task and collect interaction data.

 \noindent \textbf{Generating Task Proposals}
We follow the box-demonstration-instruction-based prompting method proposed by \cite{hong20233dllm}, and prompt ChatGPT to generate tasks.
In addition to the ground-truth bounding boxes of objects, we also input the ground-truth materials, deformability, and hardness, as well as the ground-truth temperature labels of all objects. ChatGPT is provided with a list of actions to be performed in the environment. Then it generates specific tasks requiring interactions with objects, a sequence of words representing pseudo ground-truth actions, and language reasoning outputs which are deduced from the ground-truth feedback labels of the objects (note that ChatGPT has access to all material and temperature labels, so that it could generate a sentence like ``it feels cold" after the ``touch" action). We cover a diverse set of tasks including multisensory captioning, question answering, embodied dialogue, navigation, object manipulation, tool use, rearrangement, task decomposition, and so on. We append all prompts in Supplementary Material.



 \noindent \textbf{Interaction Data Collection}
The embodied agent first randomly explores the environments to collect initial RGBD environment data. Given the actions, the agent executes the actions to interact with the objects in the environment and obtains the sensory feedback. For example, when the action is "touching an object", the agent returns the tactile and temperature readings of it. We store all the interaction results of the actions. From one interaction, we could incrementally construct several input-output data, denoting the interaction at different steps, as shown in Figure \ref{fig:dataset}.

\section{MultiPLY}
\begin{figure*}
    \centering
    \includegraphics[width=0.85\linewidth]{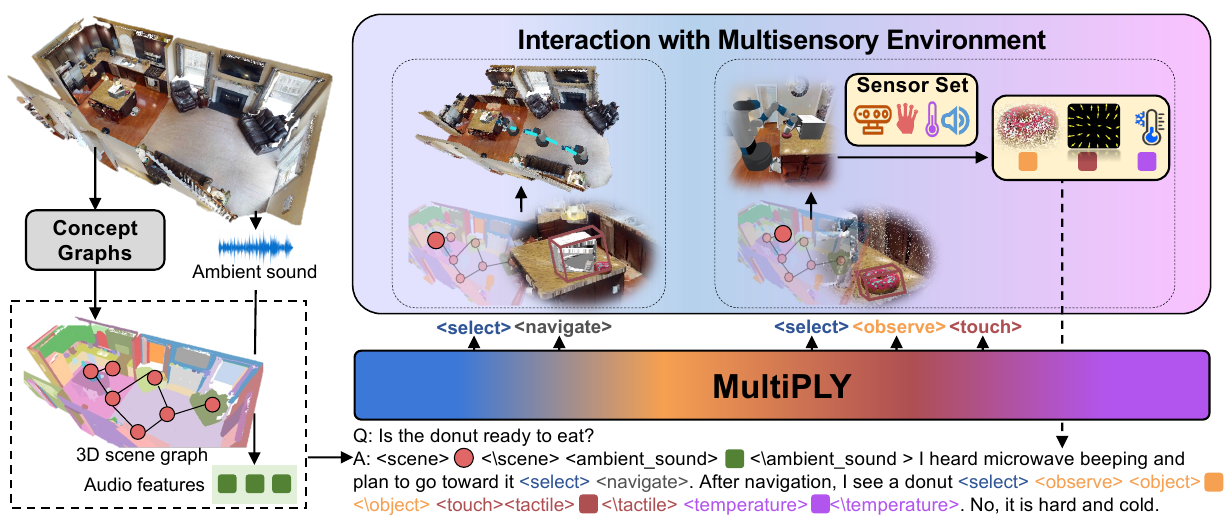}
    \vspace{-3mm}
    \caption{\textbf{Overview of our MultiPLY.} We first encode the scene as an abstracted object-centric representation, while multisensory details of objects can only be unveiled when the agent executes an action and interacts with them. We devise a set of action tokens denoting the actions of agents to interact with the environment. The interaction results are appended back to the LLM via state tokens.}
    \label{fig:framework}
\vspace{-4mm}
\end{figure*}

In this section, we introduce the MultiPLY framework. As in Figure \ref{fig:framework}, we first encode the scene as an abstracted object-centric representation, while multisensory details of objects are unveiled only when the agent executes an action and interacts with them. We devise a set of action tokens denoting the actions of agents to interact with the environment. Interaction results are appended back to the LLM via state tokens to generate subsequent text or action tokens.

\subsection{Object-Centric Scene Representations}
\label{multisensory representation}


Our model first takes the features of the 3D environment explored by the agent as inputs to form an initial impression of what the scene looks like. 
We follow 3D-LLM and utilize 2D features to construct 3D scene features, so that the visual features could be seamlessly fed into a pre-trained vision-language model without adaption. However, the point cloud encoding of 3D-LLMs makes it hard for LLMs to process thousands of points at a time. Alternatively, when humans explore a 3D environment, we abstract over the scene and roughly form an idea of objects and their locations without remembering all the details. Likewise, we propose to represent the environment as an abstracted object-centric representation. We use concept graphs \cite{gu2023conceptgraphs} powered with a CLIP \cite{radford2021learning} encoder to first encode the objects in the observed images, and fuse the outputs in images to 3D by multi-view association. We also add position embeddings to the visual features of objects. We finally get $\mathcal{O} \times 1024$ features as an abstracted object-centric scene representation, where $\mathcal{O}$ is the number of objects. If there's an ambient sound emitted by an object in the 3D environment, we encode the sound using the CLAP \cite{elizalde2022clap} audio encoder and get a $1024$-dim feature. The object-centric scene representation and ambient sound representation serve as the initial inputs to the LLM, enclosed by tokens as \texttt{<SCENE>}, \texttt{</SCENE>} and \texttt{<AMBIENT\_SOUND>}, \texttt{</AMBIENT\_SOUND>}.

\subsection{Action Tokens}
We devise a set of action tokens that denote the agent's interaction with the environment, which are listed below:
\begin{itemize}[leftmargin=0.5cm]
\item \textbf{\texttt{<SELECT>}} token selects an object to interact with. The object is chosen by the attention between the language features (\textit{i.e.}, the last hidden state of the LLM of the \texttt{SELECT} token), and the CLIP visual features of the objects in the environment. It selects the object with the maximum attention score.
\item \textbf{\texttt{<NAVIGATE>}} token asks an agent to navigate to the selected object. Note that the navigation action could be executed by any pre-defined pathfinder module and is not the research focus of this paper. 
\item \textbf{\texttt{<OBSERVE>}} token asks an agent to scrutinize an object that is chosen and get the object details (in the form of the detailed point cloud of the object).
\item \textbf{\texttt{<TOUCH>}} token allows the agent to touch the object that is chosen, to get the tactile and temperature information.
\item \textbf{\texttt{<HIT>}} token allows the agent to hit the chosen object to get the impact sound.
\item \textbf{\texttt{<PICK-UP>}}, \textbf{\texttt{<PUT-DOWN>}} tokens enable the agent to pick up or put down a chosen object.
\item \textbf{\texttt{<LOOK-AROUND>}} token allows the agent to rotate its head and get nearby objects. 

\end{itemize}

\subsection{State Tokens}
We devise another set of state tokens to feed the interaction results back to the LLM. 
\begin{itemize}[leftmargin=0.5cm]
\item \textbf{\texttt{<OBJECT>}} encodes the obtained object points when the agent \texttt{<OBSERVE>}s an object. Specifically, we get the 3D features aggregated from 2D CLIP features \cite{hong20233dllm} and add position embeddings to the 3D features. We build $\mathcal{N} \times 1024$ object point cloud features where $\mathcal{N}$ is the number of points.  
\item \textbf{\texttt{<IMPACT\_SOUND>}} encodes the obtained impact sound when the agent \texttt{<HIT>}s an object. We use CLAP audio encoder to encode the sound and get $1024$-dim impact sound representation. Since the CLAP features are not aligned with the LLM, we use a sound projector (one linear layer) to map to the feature space of the LLM. 
\item \textbf{\texttt{<TACTILE>}} encodes the obtained tactile information when an object is being \texttt{<TOUCH>}ed by an agent. We transform the tactile reading as a heatmap and use CLIP to encode the heatmap. We mean-pool over the patches and get $1024$-dim temperature features. We use a tactile projector (one linear layer) to map to the feature space of the LLM.
\item \textbf{\texttt{<TEMPERATURE>}} encodes the obtained temperature. We transform the temperature reading as a heatmap and use CLIP to encode the heatmap. We mean-pool over the patches and get $1024$-dim temperature features. We use a temperature projector (one linear layer) to map to the feature space of the LLM.
\end{itemize}

\subsection{Training \& Inference}
\noindent \textbf{Model Architecture} We use LLaVA \cite{liu2023visual} as our backbone multi-modal large language model. Since our visual features have been aligned to the same embedding space as LLaVA using ConceptGraphs \cite{gu2023conceptgraphs}, we could directly use LLaVA's vision-to-language projector without pretraining on vision-language data. For other sensor modalities, we leverage a lightweight adapter, which is a one-layer linear projector to project the sensor features into the text token embedding space of LLaVA.

\noindent \textbf{Modality Alignment} As stated above, the tactile, sound, and temperature representations are not aligned with the language features. In the first stage, we train the sensor-to-language adapter for multisensory feature alignment. For audio-language alignment, we use AudioSet \cite{Gemmeke2017AudioSA} and AudioCaps \cite{Kim2019AudioCapsGC}. For impact sound, tactile, and thermal data, we use ChatGPT to generate a one-sentence caption describing the material and the alignment between each sensor modality and language. We freeze the weight of the image encoder and the LLM for faster convergence and maintenance of language reasoning abilities. 

\noindent \textbf{Instruction tuning with Multisensory Universe} In the second stage, we tune LLaVA with our multisensory dataset. Our training loss consists of two parts. The first one is the LLM loss which is the same as the original LLaVA model. We add one more loss that forces the model to select the right object to attend to. Specifically, we calculate the attention between the last hidden state of the LLM of
the SELECT token, and each abstracted object feature. The feature goes through a Sigmoid layer, and is optimized with a binary cross entropy (BCE) loss. We unfreeze the whole model for the training of this stage. We use FSDP on 128 V100 GPUS for efficient training. 

\noindent \textbf{Inference} At the inference time, our MultiPLY first takes the task prompt and abstracted scene representation as inputs and generates subsequent tokens. Once an action token is generated, an embodied agent is instructed to take the action in Habitat-sim \cite{habitat19iccv} and interact with the environment. The observation outcome of the agent is sent back to the LLM as inputs via state tokens. The LLM further generates next tokens based on the current state inputs.
\section{Experiments}
After training on our collected Multisensory Universe, we perform an evaluation in the simulator, where an agent could actually interact with the environment when the action tokens are generated by MultiPLY. Then, the LLM waits for the agent to complete the actions and send back the observations via state tokens to generate the next token.
We provide four experimental settings: object retrieval, tool use, multisensory captioning, and task decomposition, and provide detailed task descriptions, baselines, and analysis for each task. We ensure that no scenes and objects in the Multisensory Universe appear in the evaluation setup. Due to space limits, we attach more ablative studies in the Supplementary Material, where we experiment with each possible combination of sensory inputs from different modalities, with or without interaction with the environment.
\subsection{Object Retrieval}
\textbf{Task Decription} We devise the object retrieval task where several similar objects are present in the 3D scene, and the agent needs to use multiple sensor data to retrieve the correct object. For example, the task input could be like "retrieve the soft paper cup with hot water", while there could be distracting objects like ``hard paper cup with hot water", ``soft paper cup with hot water", ``soft plastic bowl with hot water" or ``soft paper bowl with hot water", etc. The scene setup is different from the Multisensory Universe as we place more distracting objects to retrieve from (while in Multisensory Universe most scenes have two similar objects), and we include different sensor attribute combinations from Multisensory Universe objects. For example, in the training set, we saw a ceramic cup and a paper bowl, and in the evaluation, we query about a paper cup.

\noindent \textbf{Baselines} We include a set of cross-modality retrieval models as our baselines, which return the similarity between aligned sensor embeddings. They can be categorized into 1) single-sensor language models, such as CLIP and CLAP. 2) 2D multisensory models, for which the embeddings of other modalities have been mapped to the same as 2D images like ImageBind \cite{girdhar2023imagebind}.  3) 3D multisensory models, in which the embeddings of object point clouds are binded to other modalities, like PointBind \cite{guo2023pointbind}. We first explore the environment and use concept graphs to represent the scene as a set of object features like MultiPLY, where the object features are visual embeddings from these retrieval models. The select action could be achieved by calculating the similarity between the object embedding and the language embedding, and the object with the highest score will be retrieved. As these models cannot interact with the environment to get the tactile, impact sound, and temperature data, we refine three setups for the baselines: 1) No interaction, and retrieve the object with the highest retrieval score. (For CLAP we assume that we have impact sounds of all objects) 2) Interact with the environment using oracle interactive actions. That is, we first retrieve the objects of interest via visual-language similarity, then we manually control the agent to interact with the objects to get impact sound, tactile and temperature information. The embeddings of all sensors are averaged and calculate the similarities with the language query, and the object with the highest score is retrieved. Since the action tokens are pre-defined and not generated, this oracle setting makes it easier to compete with MultiPLY. 3) Finetuned with a modified version of our Multisensory Universe tailored for multi-modal alignment and retrieval. Specifically, we first align the sensor data of the objects in Multisensory Universe to visual modality (like in ImageBind and PointBind), then we further align them with the modified language data in Multisensory Universe. 

\noindent For LLM-based methods, we include Pointbind-LLM, which uses the pointbind representations and performs instruction tuning with LLaMA \cite{touvron2023llama}. We also experiment with MultiPLY-2D, a 2D variant of our model, where we replace 3D features with 2D single-view features.
\begin{table}[!ht]
    \centering
    \small
    \begin{tabular}{l|c}
    \toprule
        Model  & Retrieval Accuracy
        \\ \midrule
        ConceptGraph+CLAP  & 14.5  \\ 
        ConceptGraph+CLIP  & 18.7  \\ 
        \midrule
        ConceptGraph+ImageBind  & 20.3  \\ 
        ConceptGraph+ImageBind-I  & 24.7 \\ 
        ConceptGraph+ImageBind-I (Finetuned)  & 36.7 \\ 
        MultiPLY-2D & 44.6  \\ 
        \midrule
        ConceptGraph+PointBind  & 19.5 \\ 
        ConceptGraph+PointBind-I  & 22.7 \\ 
        ConceptGraph+PointBind-I (Finetuned) & 40.4 \\ 
        \midrule
        PointBind-LLM (Finetuned) & 48.9 \\
        \midrule
        MultiPLY & \textbf{56.7}  \\ 
        \bottomrule
    \end{tabular}
    \vspace{-3mm}
    \caption{\textbf{Experimental Results of Object Retrieval.} -I denotes the models utilize oracle action tokens to interact with the environment. (Finetuned) means finetuned on  Multisensory Universe.}
    \vspace{-3mm}
    \label{tab:retrieval}
\end{table}

\noindent \textbf{Analysis} Table \ref{tab:retrieval} shows the object retrieval results. We could come to several conclusions. First, models that take multiple sensory inputs outperform models that handle single modality inputs by a large margin. CLIP, CLAP, as well as  
models that use the initial visual embeddings have a very low score in object retrieval, emphasizing the importance of integrating multisensory data for reasoning. Second, 3D-based models surpass 2D models, mainly because single-view images sometimes fail to provide enough information to reason about the objects due to view inconsistency and occlusion. Third, LLMs outperform similarity-based retrieval models. The reason could be that retrieval models fuse the multisensory embeddings into a whole, and do not disentangle the representation, or interact with the different sensors step by step. In general, our MultiPLY outperforms the baseline models a lot. That's probably because one weakness of the binding-based methods is that they bind everything to the visual modality, while one visual attribute could be mapped to several attributes from another modality (\textit{e.g.,} from the appearance of a cup, we could not tell whether it's made of ceramic or plastic, unable to align to different impact sounds for alignment). Our MultiPLY resolves ambiguity by interacting with and reasoning about the different sensor data individually.

\subsection{Tool Use}
\noindent \textbf{Task Description} 
In an embodied environment, multisensory data are crucial for finding an appropriate tool to solve a problem. One example is that when we are injured, we need to retrieve warm compresses or ice packs depending on the injured parts and how long we've been injured. We could also find substitute tools if the common ones are not present. For example, we could use a steel spoon to replace the can opener, but we can't use a plastic spoon. Similar to the object retrieval task, we place some objects from different categories, and also objects from the same categories but with different materials/haptic/thermal information in the environment. We use one sentence to describe the current situation and the goal to be done, and ask the agent to retrieve the correct tool for dealing with the situation.

\noindent \textbf{Baselines} We use the same baselines as the object retrieval experiment for tool retrieval. For LLM-based methods, we also need to give reasons when we select the tools. 

\noindent \textbf{Analysis} Table \ref{tab:tool} shows the results of tool use. We could see that the binding-based methods have a very poor performance in tool use. It might be because that they treat the object sensory data as a whole, unable to disentangle the individual sensory information such as material from the representation, let alone reasoning about how this property could be utilized as a tool, and how to analyze and deduce the functionality of an object when the multisensory information is integrated. 
\begin{table}[!ht]
    \centering
    \small
        \vspace{-3mm}

    \begin{tabular}{l|c}
    \toprule
        Model  & Accuracy
        \\ \midrule
        ConceptGraph+CLIP  & 10.1  \\ 
        \midrule
        ConceptGraph+ImageBind  &  7.4 \\ 
        ConceptGraph+ImageBind-I  & 8.2 \\ 
        ConceptGraph+ImageBind-I (Finetuned)  & 16.4 \\ 
        MultiPLY-2D & 36.3  \\ 
        \midrule
        ConceptGraph+PointBind  & 11.5  \\ 
        ConceptGraph+PointBind-I  & 13.2 \\ 
        ConceptGraph+PointBind-I (Finetuned) & 18.7 \\
        \midrule
        PointBind-LLM (Finetuned) & 32.1 \\
        \midrule
        MultiPLY  & \textbf{41.6}  \\ 
        \bottomrule
    \end{tabular}
    \vspace{-3mm}
    \caption{\textbf{Experimental Results of Tool Use.}}
    \vspace{-3mm}
    \label{tab:tool}
\end{table}

\subsection{Multisensory Captioning}
\noindent \textbf{Task Description} Different from traditional single-modality captioning tasks, multisensory captioning requires the model to describe the object in all senses. By giving semantic information about an object or ambient sound emitted by the object, the agent must first navigate to the object to interact with it and describe it.

\begin{figure*}[t]
    \centering
    \includegraphics[width=0.85\textwidth, 
        trim={0cm, 3.85cm, 0cm, 0.15cm}, clip]{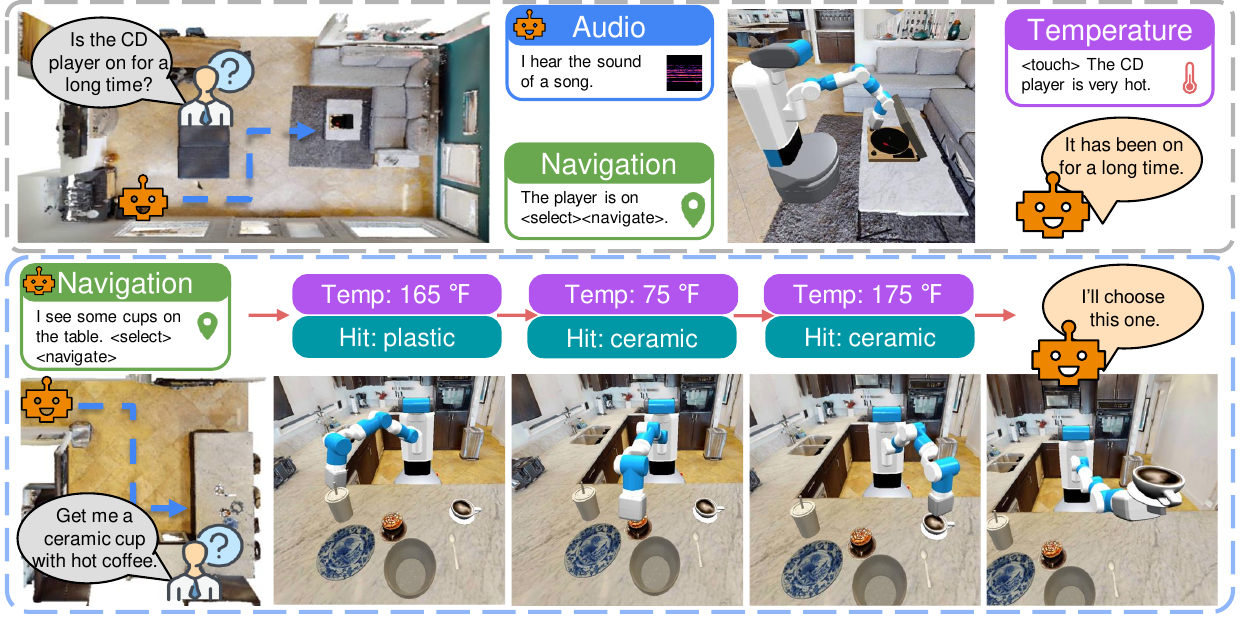}
    \caption{\textbf{Qualitative Examples of our MultiPLY}. MultiPLY could interact with the objects in the embodied environments and gather multisensory information.}
    \label{fig:qualitative}
    \vspace{-3mm}
\end{figure*}

\noindent \textbf{Baselines} For baseline models, we include LLaVA, which takes a holistic scene image as input and generates a caption about the queried object. 3D-LLM takes the scene point cloud as inputs, and uses dense captioning to describe the object. Both methods only use visual information. PointBind-LLM  first retrieves the objects by modality alignment, and then interacts with the objects and integrates multisensory information to describe the queried object. 
\begin{table}[!ht]
    \centering
    \small
        \vspace{-3mm}

    \begin{tabular}{l|ccc}
    \toprule
        ~ & BLEU1 & BLEU4 & METEOR \\ \midrule
        LLaVA &9.5& 0.6& 7.1\\
        LLaVA (Finetuned) &28.6 &10.1 & 10.4\\
        3D-LLM & 14.4 & 1.5 & 9.5  \\
        3D-LLM (Finetuned) & 31.2 & 12.1 & 12.4 \\        
        PointBind-LLM & 16.5 & 2.3 & 7.7  \\ 
        PointBind-LLM (Finetuned) & 36.7 & 14.5 & 15.1  \\ 
        \midrule
        MultiPLY & \textbf{48.9} & \textbf{20.1} & \textbf{24.2}   \\ 
        \bottomrule
    \end{tabular}
        \vspace{-3mm}

    \caption{\textbf{Experimental Results of Multisensory Captioning.}}
        \vspace{-5mm}

    \label{tab:captioning}
\end{table}

\noindent \textbf{Analysis} Table \ref{tab:captioning} shows the result. From the table, we could see that 3D-based LLMs overall outshine 2D VLMs. LLaVA and 3D-LLM take the holistic representation as inputs, and thus fail to compete with models that could interact with the models to switch between representations. MultiPLY outshines Pointbind-LLM, probably because PointBind binds the representations of different modalities, making it difficult to disentangle the senses.
\subsection{Task Decomposition}
\noindent \textbf{Task Definition} Task decomposition focuses on decomposing a high-level task into smaller actions. In our setting, we focus on retrieving different things to prepare for a task. For example, to prepare for dinner, we need to first detect available foods in the kitchen, and gauge its temperature. If it's cold, we need to heat it in the microwave so we also need to retrieve a ceramic or glass container which is microwave-safe. We also need to prepare the utensils of the appropriate materials. In our setting, we place several possible choice combinations in the environment, we also place object combinations unseen from the Multisensory Universe. As long as the agent retrieves one of the correct combinations, the task is marked as success.

\noindent \textbf{Baselines} We include LLaVA, a minimal 2D image version of our model. We output an image of the scene and ask the model to decompose the tasks into actions. We also utilize 3D-LLM since it's capable of performing task decomposition. In the original paper, we take the whole point cloud as input and generate low-level actions. Note that there is a domain gap between the task decomposition data 3D-LLM was trained on and our setting, which yields almost zero success rates of 3D-LLM without finetuning. Therefore, we finetune all models as baselines. For each baseline we have two variants: 1) wo Interaction: generate all actions all at once, and execute the actions sequentially in the environment; 2) w Interaction: generate an action one at a time, take the action feedback and generate the next action.

\begin{table}[!ht]
    \centering
        \vspace{-3mm}

    \begin{tabular}{l|ccc}
    \toprule
        ~ & success rate \\ \midrule
        LLaVA wo Interaction & 4.0 \\
        LLaVA w Interaction & 14.5 \\
        3D-LLM wo Interaction & 8.7 \\
        3D-LLM w Interaction & 22.4 \\        
        \midrule
        MultiPLY & \textbf{30.2}   \\ 
        \bottomrule
    \end{tabular}
        \vspace{-3mm}

    \caption{\textbf{Experimental Results of Multisensory Captioning.}}
        \vspace{-3mm}

    \label{tab:task}
\end{table}

\noindent \textbf{Analysis} Table \ref{tab:task} shows the task decomposition results. From the table, we observe that models without interaction have very poor results, probably because vision-language models have hallucination to a great extent. For example, the models could generate ``retrieve a bread" when there's no bread in the scene. MultiPLY outperforms the baseline models by a large margin. One reason could be that MultiPLY leverages multisensory information while the other two leverage visual information. The other reason might be that baseline models take the whole scene as inputs, thus could not attend to the nuanced object in the scene.

\subsection{Qualitative Examples}
Qualitative Examples are shown in Figure \ref{fig:qualitative}, demonstrating the power of MultiPLY to interact with objects in the embodied environments and gather multisensory information. More examples can be found in the \textbf{supplementary materials}.

\section{Conclusion}
In this paper, we propose MultiPLY, a multisensory LLM that could incorporate multisensory interactive data into large language models. We introduce Multisensory Universe, a dataset comprising 500k multisensory data collected by an agent actively exploring and interacting with an environment. One limitation of our model is that currently MultiPLY does not involve detailed navigation and control policy, but utilizes pre-defined policies for carrying out the actions. We think that such aspects are orthogonal to our study, and could be explored and seamlessly integrated into our framework in the future. 

{
    \small
    \bibliographystyle{ieeenat_fullname}
    \bibliography{main}
}

\clearpage
\onecolumn
\appendix

{
    \hypersetup{linkcolor=black}
    \tableofcontents
}
\clearpage

\section{Dataset}
\subsection{More details on Scene Construction}
\begin{figure*}[htbp]
    \centering
    \includegraphics[width=\linewidth]{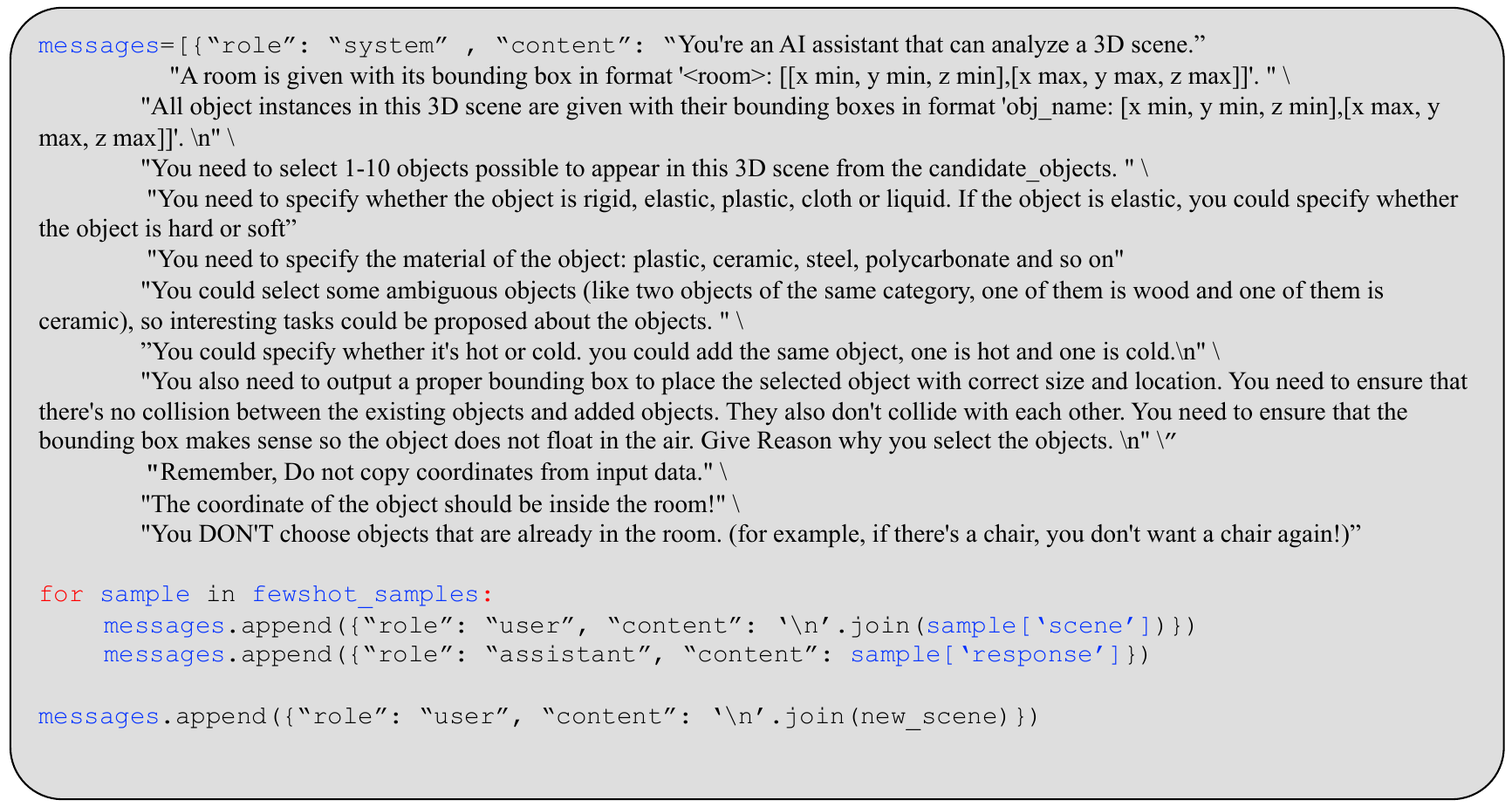}
    \caption{Prompts for adding objects to the scene}
    \label{fig:scene}
    \vspace{-3mm}
\end{figure*}

In figure \ref{fig:scene}, we show how we add new objects to the HM3D scenes. Specifically, ChatGPT is asked to generate: 1) object bounding boxes; 2) object material and material properties; 3) temperatures.

\subsection{More details on Sensor Data Acquisition}
In this section, we elaborate on how we get the sensor data of the objects in details.
\subsubsection{Tactile}
DiffTactile \cite{anonymous2023difftactile} requires us to provide a set of parameters for tactile simulation of different objects. In addition to telling to the model whether we are inputting a rigid, elastic, or elasto-plastic object, we also need to specify the parameters such as Young's modulus, Poisson's ratio, Yield Strength and so on. 

\begin{figure*}[htbp]
    \centering
    \includegraphics[width=\linewidth]{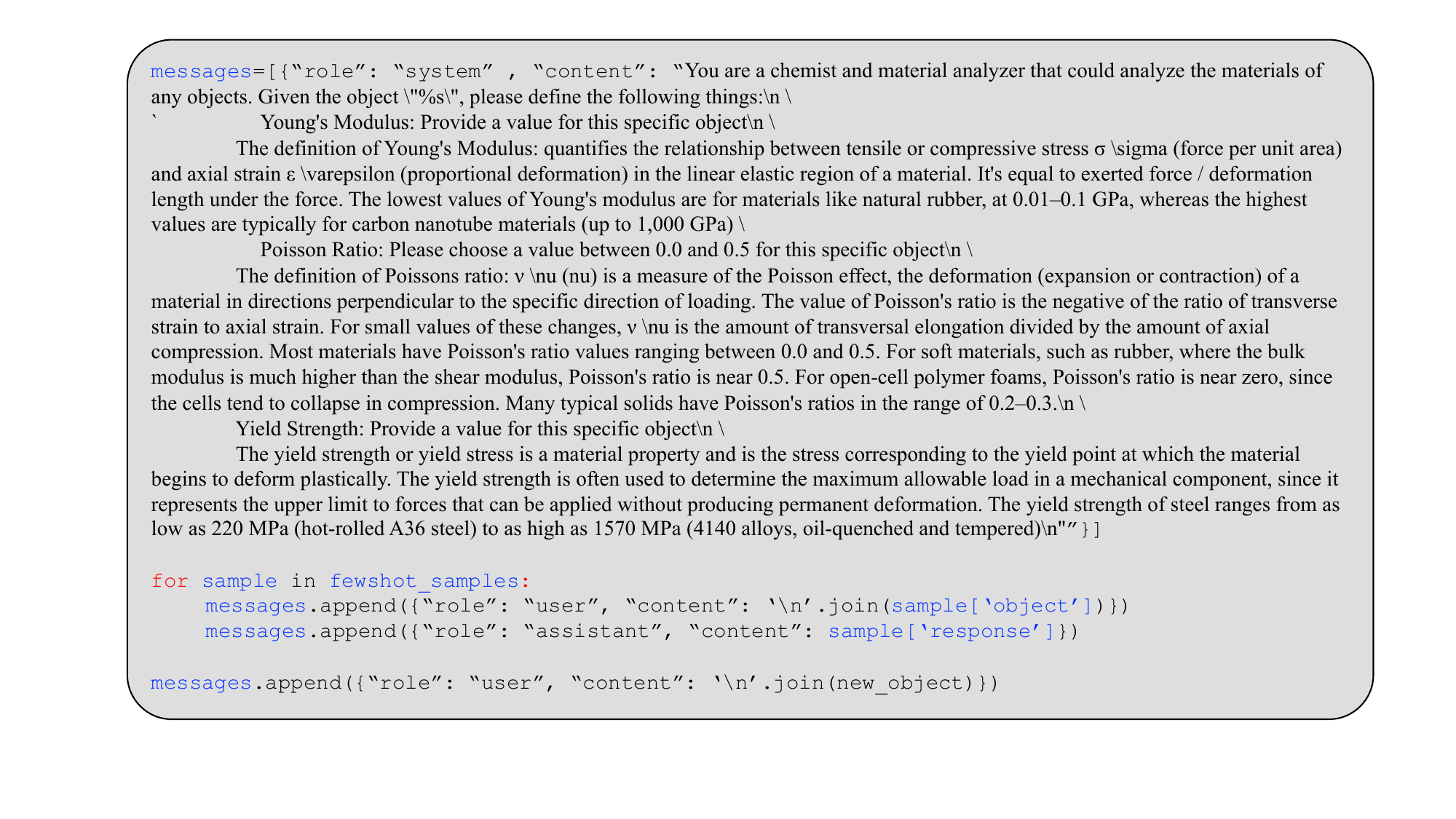}
    \caption{Prompts for getting the material parameters for the objects}
    \label{fig:material}
    \vspace{-3mm}
\end{figure*}

As in the main paper, when ChatGPT adds objects to the scene, it also specifies what kinds of objects (\textit{e.g}, rigid, elastic, plastic) and the softness / deformability (in the description of language) of each object. In order to get the parameters required by DiffTactile, we prompt ChatGPT with the type and the softness / deformability description, as well as detailed definition of each parameter, and the possible values of the parameters of several few-shot examples. ChatGPT is asked to return the detailed parameter combinations of the given objects. For example, a soft bread corresponds to a smaller young's modulus, while a harder one corresponds to a larger young's modulus. We add the prompt in getting the parameters in Figure \ref{fig:material}.

\begin{figure}[htp]
\centering
\includegraphics[width=.3\textwidth]{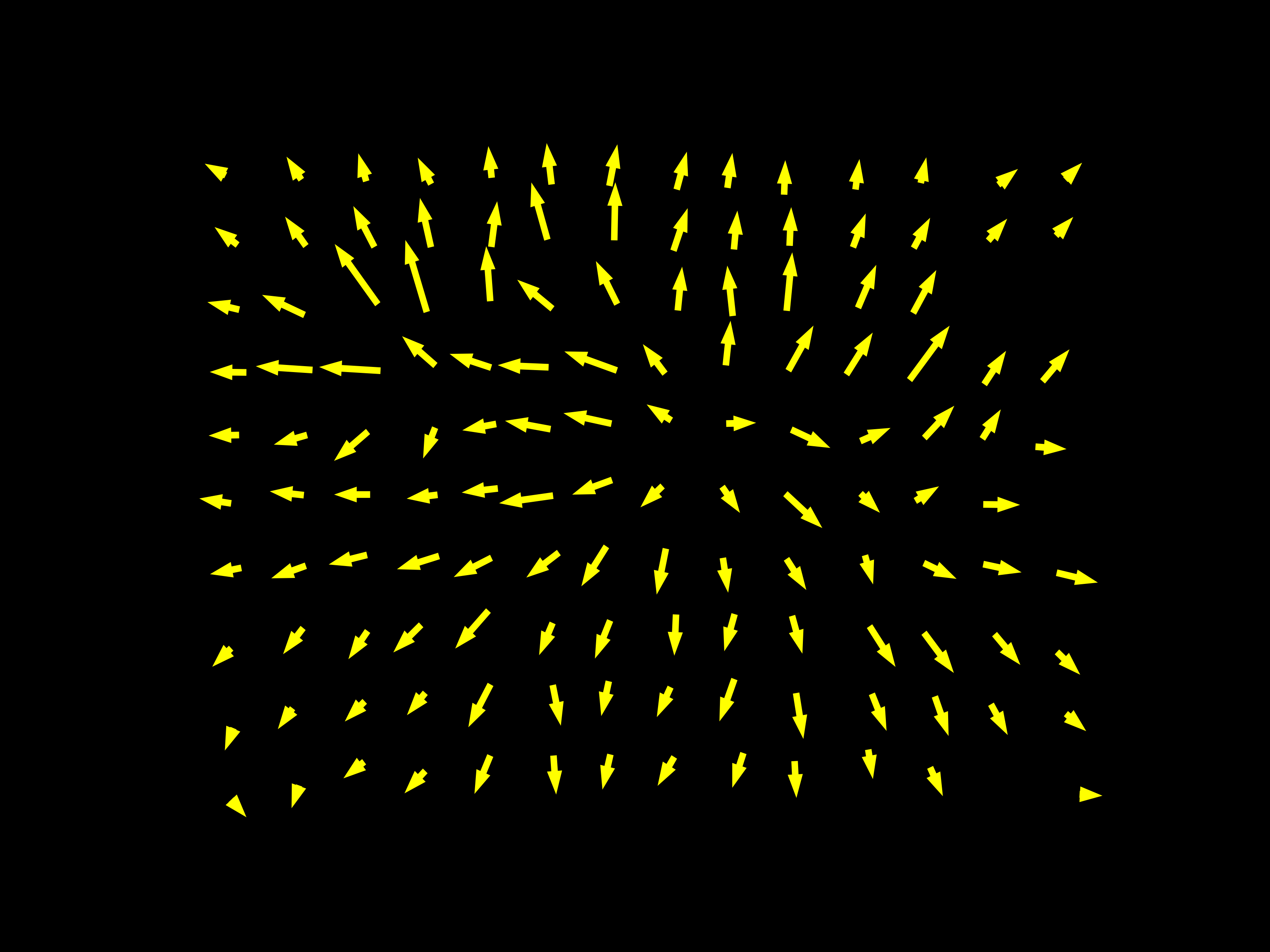}\hfill
\includegraphics[width=.3\textwidth]{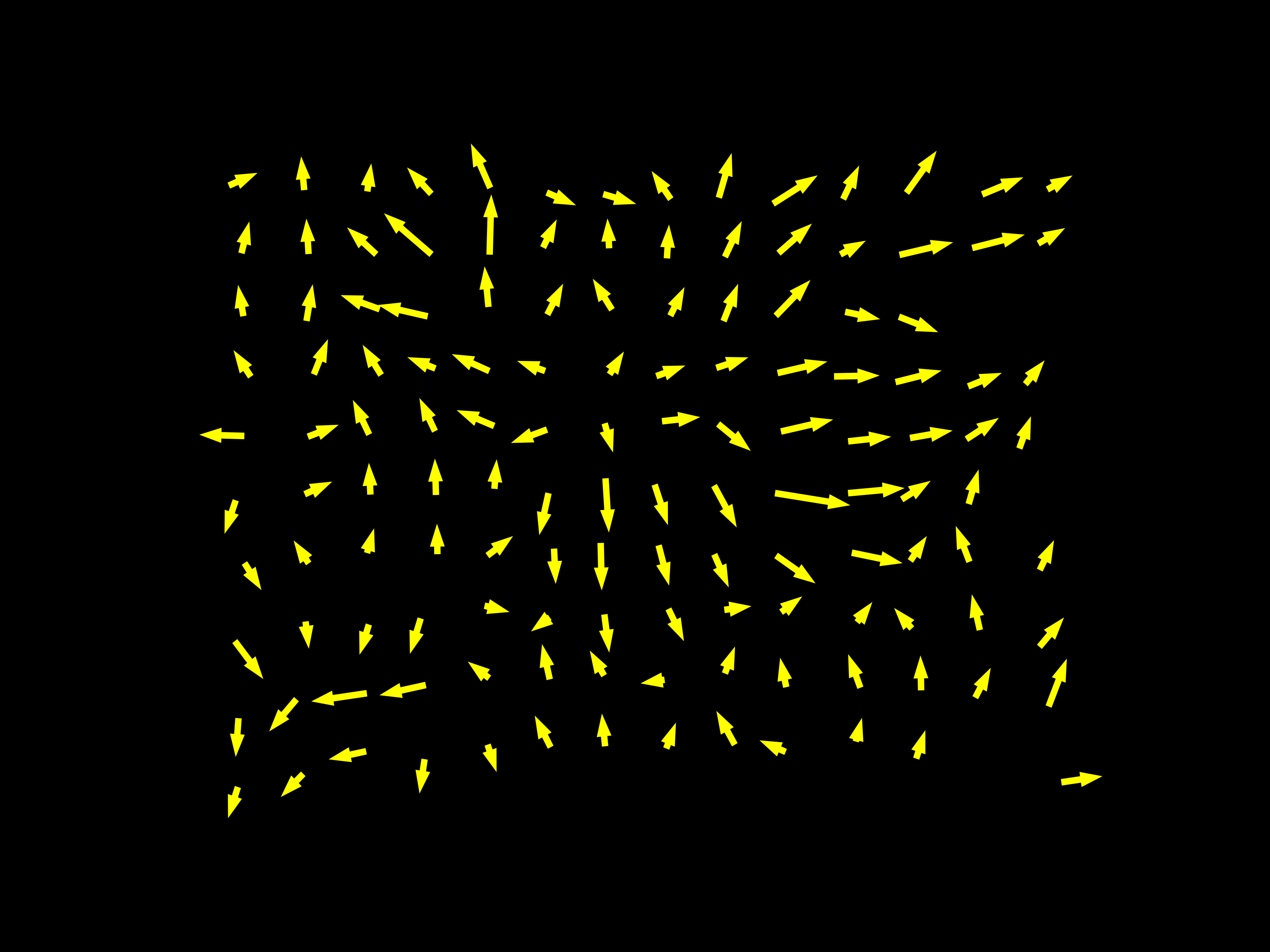}\hfill
\includegraphics[width=.3\textwidth]{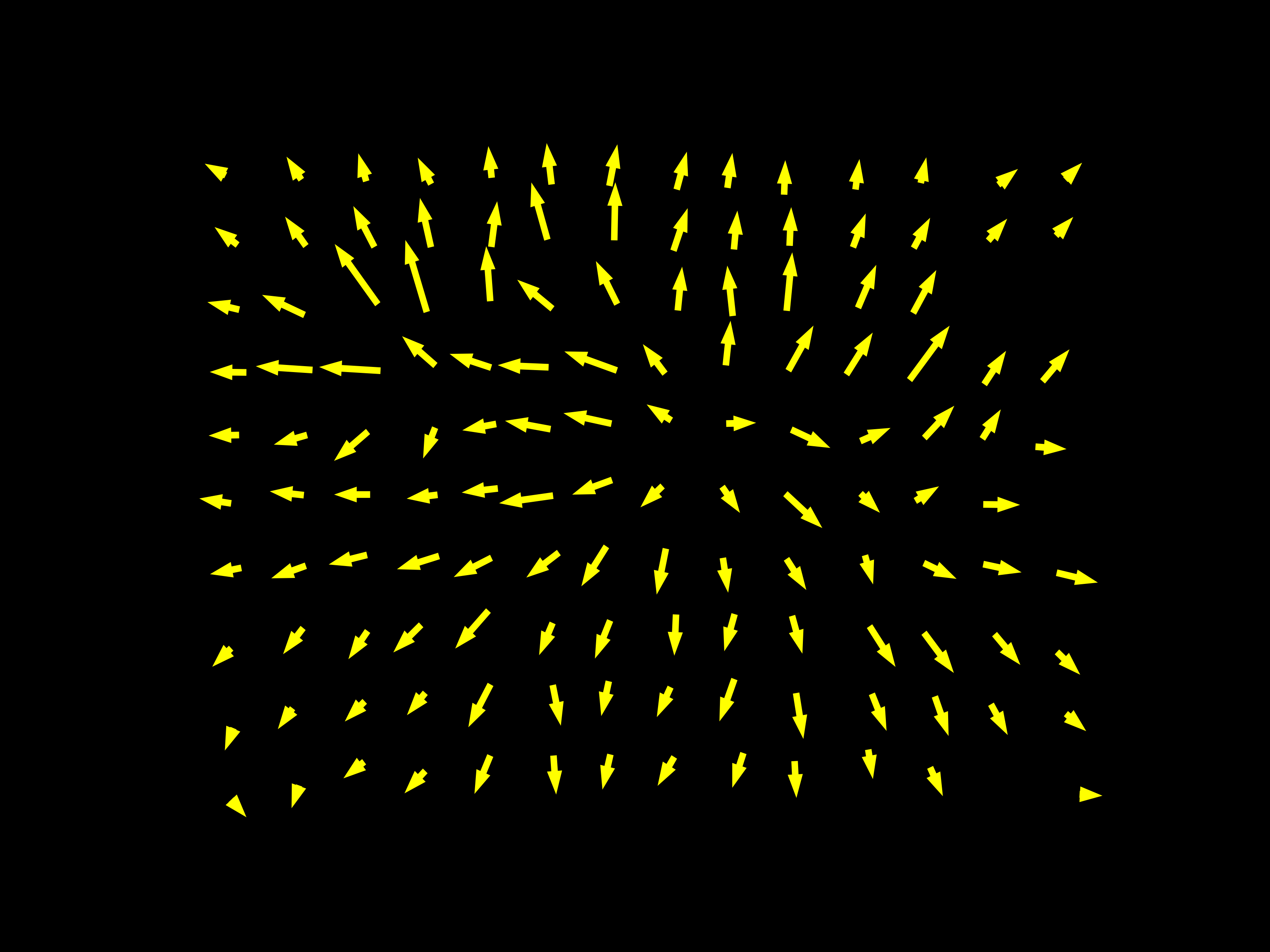}

\caption{Examples of tactile images.}
\label{fig:tactile}
\end{figure}

We input the object into DiffTactile, normalize the shape of the gripper according to the object. We record the 2D initial position and final position of the markers in the gripper. And we turn the tactile readings into a 2D image, by drawing an arrowed line from the initial position to the final position. We show some examples of tactile images in Figure \ref{fig:tactile}. We sample 16 touching positions of each object. In training and evaluation, we randomly return one image of the object.

\subsubsection{Impact Sound}
ObjectFolder \cite{Gao2021ObjectFolderAD} stores the multi-modal information all in implicit fields. That is, by inputting a striking location to the sound implicit field of an object, we could get the impact sound of striking the object at the specific location. For each object, we randomly sample 10 locations in the mesh points to get the impact sound.  In training and evaluation, we randomly return one impact sound of the object.

\subsubsection{Ambient Sound}
\begin{figure*}[htbp]
    \centering
    \includegraphics[width=\linewidth]{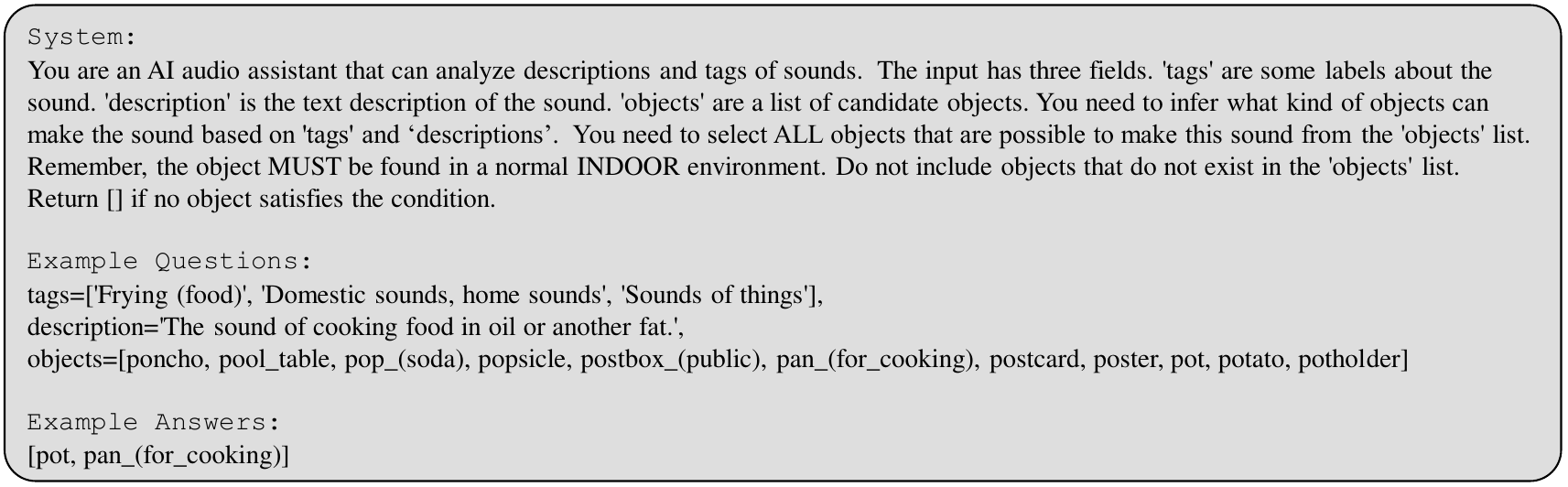}
    \caption{Prompts to match AudioSet with Objects}
    \label{fig:supp_sound}
    \vspace{-3mm}
\end{figure*}

AudioSet is paired with objects to represent ambient sound. The AudioSet ontology is organized in a hierarchy structure. From the root node to the leaf node, the description granularity becomes finer (e.g., Music - Musical instrument - Keyboard - Piano - Electric piano). Each ontology entry is attached with a description (e.g., ``Glass: sounds associated with the non-crystalline amorphous solid that is often transparent and has widespread practical, technological, and decorative uses"). Each audio is labeled with multiple ontology entries tracing from the child node to the root node (e.g., the sound of the piano will be labeled with ``Piano", ``Keyboard", ``Musical Instrument", and ``Music", but without  ``Electric piano" since this piano is not electric). We prompt ChatGPT to match each ontology entry with object categories (Figure \ref{fig:supp_sound}). 

\subsubsection{Temperature}
\begin{figure*}[htbp]
    \centering
    \includegraphics[width=\linewidth]{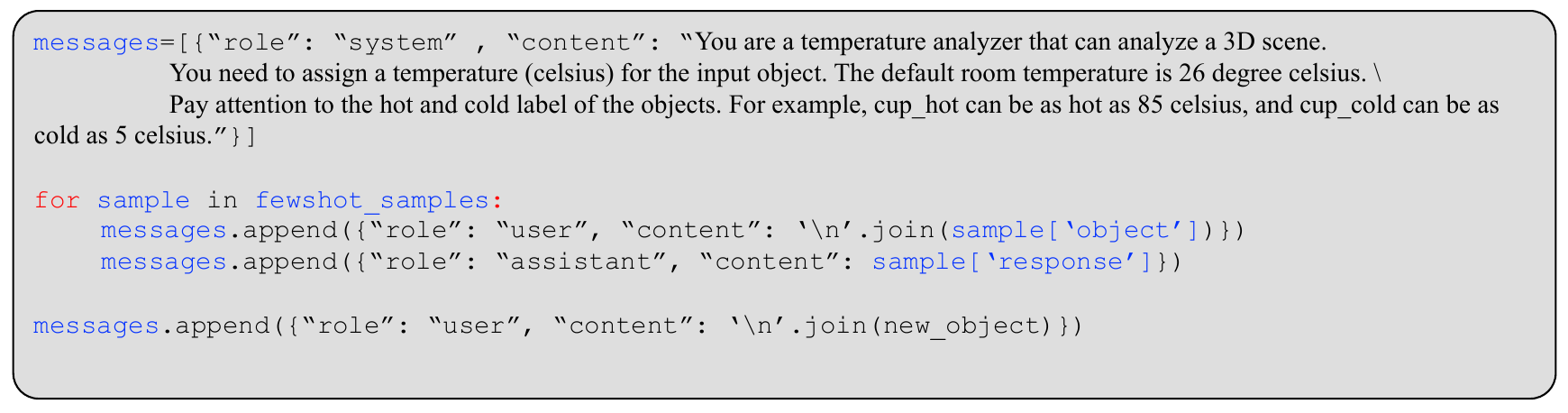}
    \caption{Prompts on generating temperature for each object}
    \label{fig:temp}
    \vspace{-3mm}
\end{figure*}

We add the prompt in getting the temperature in Figure \ref{fig:temp}.

\subsection{More details on Task Construction}

\begin{figure*}[htbp]
    \centering
    \includegraphics[width=\linewidth]{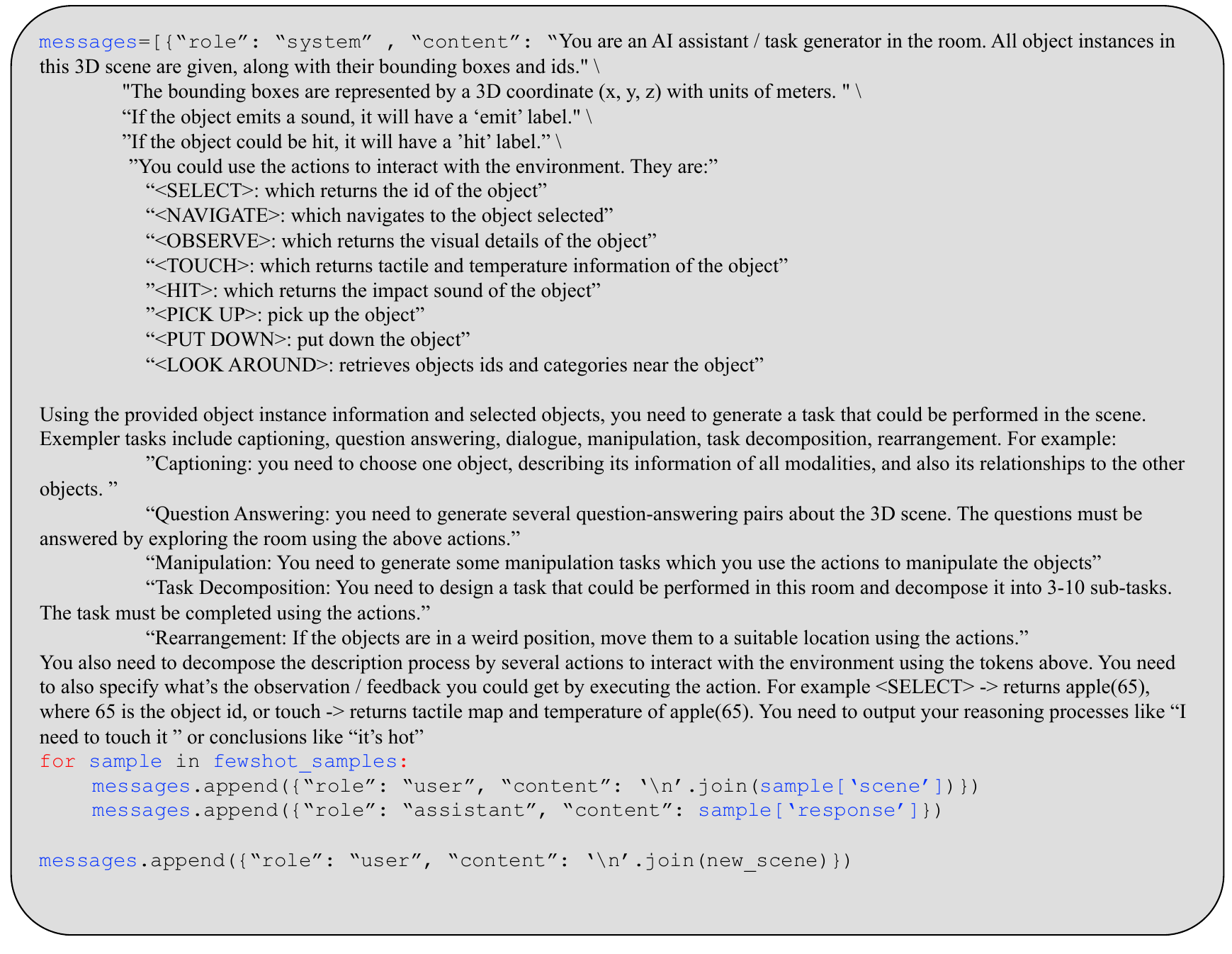}
    \caption{Prompts for task construction}
    \label{fig:task}
\end{figure*}

In Figure \ref{fig:task}, we illustrate the prompts for generating the language task data for Multisensory-Universe. Specifically, the actions could return the expected observation in the form of language (\textit{e.g.,} tactile map of and object when touching). We insert that into the state tokens for placeholder, and after the agent has executed the actions in the space and gets the observations, we append the observations back to the state tokens.

\begin{table}[htbp]
    \centering
    \begin{tabular}{l|c}
        \toprule
        Ablative Model & Acc \\ \hline
        MultiPLY Vision & 21.0 \\ 
        MultiPLY Audio & 13.2 \\ 
        MultiPLY Tactile & 10.5 \\ 
        MultiPLY Temperature & 11.2 \\ 
        \hline
        MultiPLY Vision, Audio & 31.8 \\ 
        MultiPLY Vision, Tactile & 24.3 \\ 
        MultiPLY Vision, Temperature & 25.7 \\ 
        MultiPLY Audio, Tactile & 20.6 \\ 
        MultiPLY Audio, Temperature & 23.4 \\ 
        MultiPLY Tactile, Temperature & 18.9 \\ 
        \hline
        MultiPLY Vision, Audio, Tactile & 45.3 \\ 
        MultiPLY Vision, Tactile, Temperature & 41.4 \\ 
        MultiPLY Vision, Audio, Temperature & 45.3 \\ 
        MultiPLY Audio, Tactile, Temperature & 37.7 \\ 
        \hline 
        MultiPLY & 56.7 \\
        \bottomrule
    \end{tabular}
    \caption{Ablative Study of MultiPLY}
    \label{tab:ablative}
\end{table}

\section{Experiments}
\subsection{Experimental Details}
We tune the model based on the llava-v1.5-7b checkpoint of the LLaVA model. We use Adam optimizer with learning rate of 1e-6. We train the model on 4*132 V100s. We use a batch size of 2112. The training of multi-modal adapters takes 2 hours, while the whole finetuning takes less day 1 day to complete.

 We use the mm projector of the original LLaVA for adapting scene representations and object point clouds to the LLM. The sound, tactile and temperature adapters are all one linear layer with input size 1024 and output size 1024. 

We use the default CLIP vision encoder of LLaVA to encode all objects, point clouds, tactile and temperature images. Specifically, for objects, we use segment anything \cite{Kirillov2023SegmentA} to get the objects out of 2D objects, mask out other objects and background, and crop the image to the size of the object, and use CLIP encoder to encode the object. We follow ConceptGraph \cite{gu2023conceptgraphs} to merge the objects from 2D to 3D. For scene construction, each object has one CLIP feature. For object details (point cloud), we project the 2D pixels of the objects to 3D, and get the point clouds of the objects.

\subsection{Ablative Studies}
In Table \ref{tab:ablative}, we show additional experimental results where we explore MultiPLY with single, double or triple modalities.

\subsection{More Qualitative Examples}

\begin{figure*}[htbp]
    \centering
    \includegraphics[width=\linewidth]{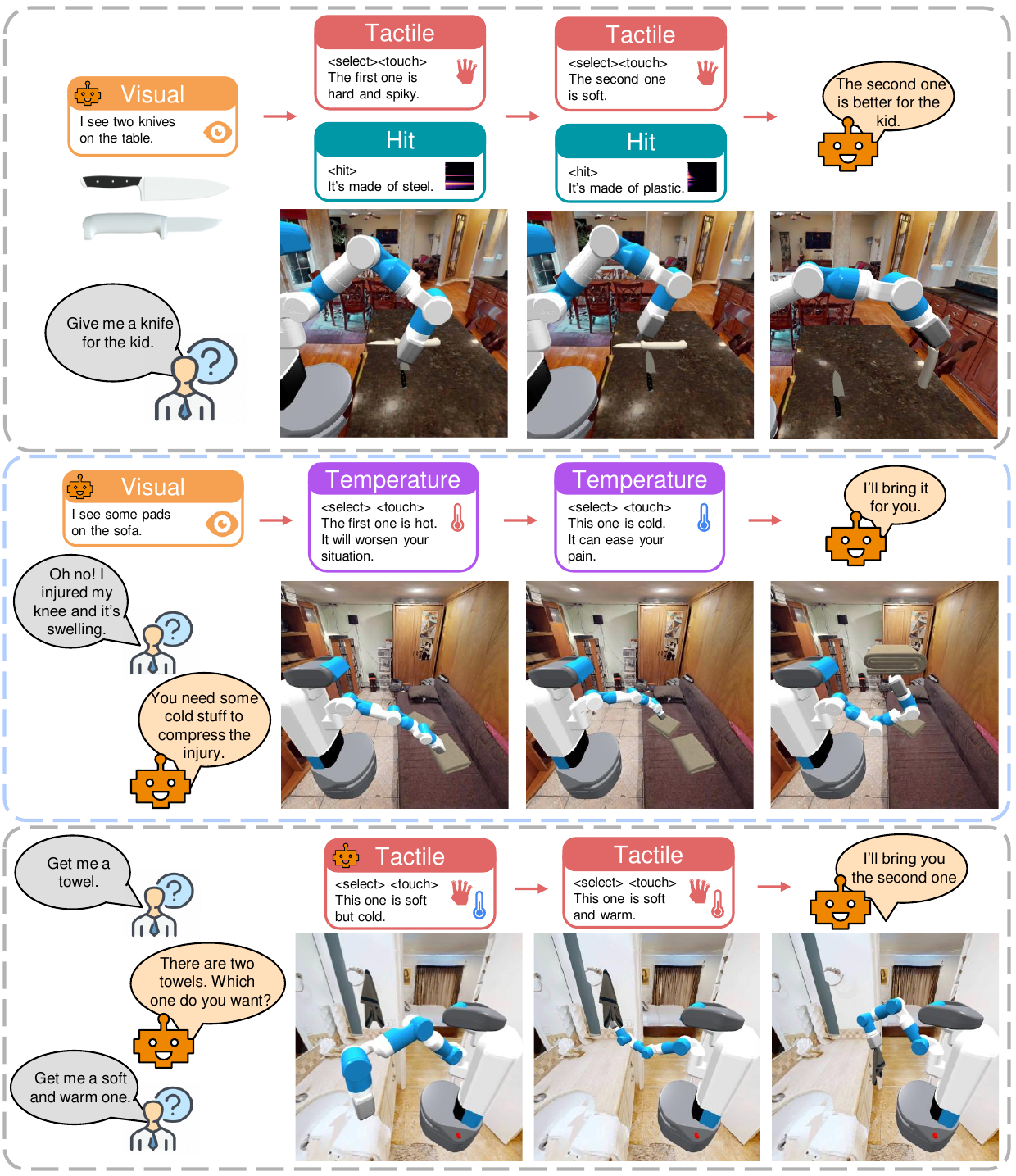}
    \caption{More qualitative examples of MultiPLY}
    \label{fig:supp_qualitative}
\end{figure*}

\end{document}